\documentclass[a4paper,twoside,11pt]{article}

\usepackage[USenglish]{babel} 
\usepackage[T1]{fontenc}

\usepackage{graphicx}
\usepackage[labelsep=endash, figurename=Figure, tablename=Table, textfont=it]{caption}
\usepackage{subfigure}
\usepackage{float}  
\usepackage{hyperref, wasysym}
\usepackage{multicol}
\usepackage{comment}

\usepackage{fancyhdr}
\usepackage[top=2.5cm, bottom=2.5cm, left=2.5cm, right=2.5cm]{geometry} 
\usepackage[usenames,dvipsnames]{pstricks}
\usepackage[round,comma,authoryear]{natbib}
\usepackage{titling}

\usepackage{amsmath,amsthm}
\usepackage{amssymb}
\usepackage{amsfonts}
\usepackage{nicefrac}       
\usepackage{graphicx}
\usepackage{microtype}
\usepackage{bbm}
\usepackage{mathrsfs}
\usepackage{fancyhdr}
\usepackage[T1]{fontenc}

\definecolor{darkblue}{rgb}{0,0,0.5}
\hypersetup{colorlinks=true,linkcolor=black,citecolor=darkblue,urlcolor=darkblue} 

\makeatletter
\let\ps@plain\ps@fancy
\makeatother

\begin{document}


\title{\bf Generalizable Implicit Neural Representation As a Universal Spatiotemporal Traffic Data Learner}
\author{Tong Nie$^{a, b}$, Guoyang Qin$^{a}$, Wei Ma$^{b, *}$ and Jian Sun$^{a, *}$}
\date{\today}

\pretitle{\centering\Large}
\posttitle{\par\vspace{1ex}}

\preauthor{\centering}
\postauthor{\par\vspace{1ex}
$^{a}$ Department of Traffic Engineering, Tongji University, Shanghai, China\\
nietong@tongji.edu.cn,
2015qgy@tongji.edu.cn,
sunjian@tongji.edu.cn\\
$^{b}$ Department of Civil and Environmental Engineering, The Hong Kong Polytechnic University, Hong Kong SAR, China\\
wei.w.ma@polyu.edu.hk\\
$^{*}$ Corresponding authors

\vspace{1ex}\it
Extended abstract accepted for presentation at the Conference in Emerging Technologies in Transportation Systems (TRC-30)\\
September 2-3, 2024, Crete, Greece\\
\vspace{1ex}

}

\maketitle
\vspace{-1cm}
\noindent\rule{\textwidth}{0.5pt}\vspace{0cm}

\fancypagestyle{firststyle}{
\lhead[]{}
\rhead[]{}
\lfoot[TRC-30]{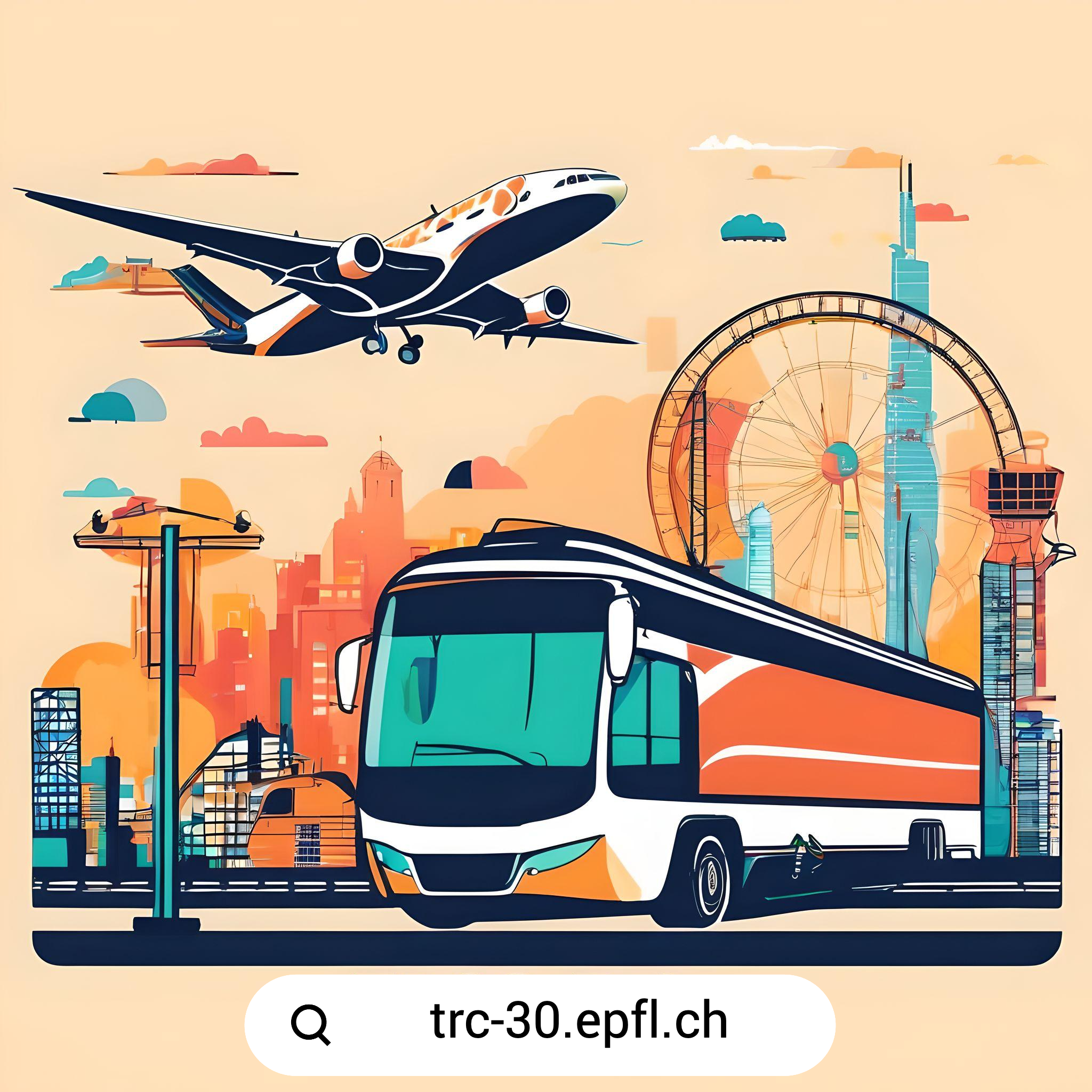}
\rfoot[Original abstract accepted for presentation]{Original abstract accepted for presentation}
\cfoot[]{}
}
\thispagestyle{firststyle}

\pagestyle{fancy}
\fancyhead{}
\fancyfoot{}
\renewcommand{\headrulewidth}{0pt}
\renewcommand{\footrulewidth}{0pt}
\setlength{\headheight}{15pt}
\lhead{Nie, Qin, Ma$^{*}$, and Sun$^{*}$}
\rhead[\thepage]{\thepage}
\lfoot[TRC-30]{TRC-30}
\rfoot[Original abstract accepted for presentation]{Original abstract accepted for presentation}
\cfoot[]{}


\section{INTRODUCTION}\label{Introduction}

The unpredictable elements involved in a vehicular traffic system, such as human behavior, weather conditions, energy supply and social economics, lead to a complex and high-dimensional dynamical transportation system. 
To better understand this system, Spatiotemporal Traffic Data (STTD) is often collected to describe its evolution over space and time. This data includes various sources such as vehicle trajectories, sensor-based time series, and dynamic mobility flow. The primary aim of STTD learning is to develop data-centric models that accurately depict traffic dynamics and can predict complex system behaviors.

\begin{figure}[!htb]
\centering
\includegraphics[width=1\textwidth]{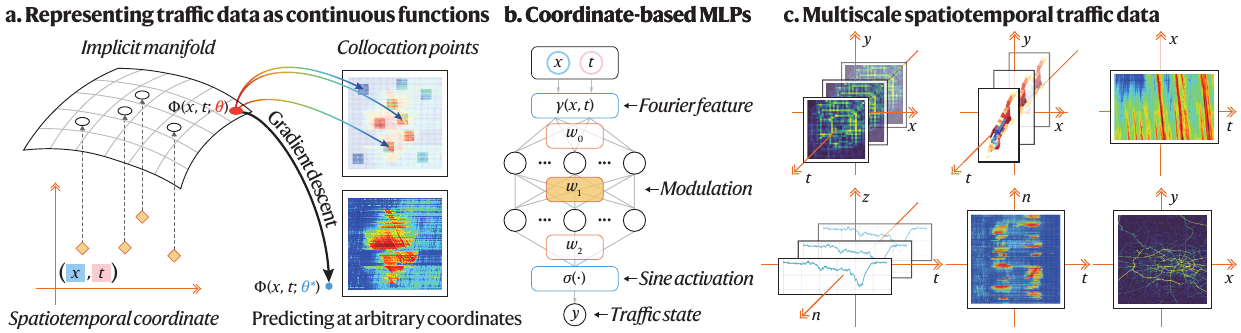}
\caption{\textbf{Representing spatiotemporal traffic data as implicit neural functions.} \textbf{(a)} Traffic data at arbitrary spatiotemporal 
 coordinates can be represented as a continuous function in an implicit space. \textbf{(b)} Coordinate-based MLPs map coordinates to traffic states. \textbf{(c)} With the resolution-independent property, our model can represent various spatiotemporal traffic data.}
\label{fig:intro}
\end{figure}

Despite its complexity, recent advances in STTD learning have found that the dynamics of the system evolve with some dominating patterns and can be captured by some low-dimensional structures.
Notably, low-rankness is a widely studied pattern, and models based on it assist in reconstructing sparse data, detecting anomalies, revealing patterns, and predicting unknown system states. 
However, these models have two primary limitations:
1) they often require a grid-based input with fixed spatiotemporal dimensions, restricting them from accommodating varying spatial resolutions or temporal lengths; 2) the low-rank pattern modeling, fixed on one data source, may not generalize to different data sources. For instance, patterns identified in one data type, such as vehicle trajectories, may not be applicable to differently structured data, such as OD demand.
These constraints mean that current STTD learning depends on data structures and sources. This limits the potential for a unified representation and emphasizes the need for a universally applicable method to link various types of STTD learning.


To address these limitations, we employ a novel technique called implicit neural representations (INRs) to learn the underlying dynamics of STTD. INRs use deep neural networks to discern patterns from continuous input \citep{SIREN,FourierFeature}. They function in a continuous space and take domain coordinates as input, predicting the corresponding quantity at queried coordinates.
INRs learn patterns in implicit manifolds and fit processes that generate target data with functional representation.
This differentiates them from low-rank models that depend on explicit patterns, enhancing their expressivity, and enabling them to learn dynamics implicitly. Consequently, they eliminate the need for fixed data dimensions and can adjust to traffic data of any scale or resolution, allowing us to model various STTD with a unified input.
In this work, we exploit the advances of INRs and tailor them to incorporate the characteristics of STTD, resulting in a novel method that serves as a universal traffic data learner (refer to Fig. \ref{fig:intro}).

Our proof-of-concept has shown promising results through extensive testing using real-world data. The method is versatile, working across different scales - from corridor-level to network-level applications. It can also be generalized to various input dimensions, data domains, output resolutions, and network topologies. This study offers novel perspectives on STTD modeling and provides an extensive analysis of practical applications, contributing to the state-of-the-art.
To our knowledge, this is the first time that INRs have been applied to STTD learning and have demonstrated effectiveness in a variety of real-world tasks. 
We anticipate this could form the basis for developing foundational models for STTD.

\section{METHODOLOGY}
To formalize a universal data learner, we let MLPs be the parameterization $\theta$.
Concretely, the function representation is expressed as a continuous mapping from the input domain to the traffic state of interest: $\Phi_{\theta}(x,t):\mathcal{X}\times\mathcal{T}\mapsto\mathcal{Y}$, where $\mathcal{X}\subseteq\mathbb{R}^{N}$ is the spatial domain, $\mathcal{T}\subseteq\mathbb{R}^{+}$ is the temporal domain, and $\mathcal{Y}\subseteq\mathbb{R}$ is the output domain. $\Phi_{\theta}$ is a coordinate-based MLP (Fig. \ref{fig:intro} (b)).

\subsection{Encoding high-frequency components in function representation}
High-frequency components can encode complex details about STTD. 
To alleviate the spectral bias of neural network towards low-frequency patterns, we adopt two advanced techniques to enable $\Phi_{\theta}$ to learn high-frequency components. Given the spatial-temporal input coordinate $\mathbf{v}=(x,t)\subseteq\mathbb{R}\times\mathbb{R}^+$, the frequency-enhanced MLP can be formulated as:
\begin{equation}\label{eq:freqmlp}
    \mathbf{h}^{(1)} = \texttt{ReLU}(\mathbf{W}^{(0)}\gamma(\mathbf{v})+\mathbf{b}^{(0)}), 
    ~\mathbf{h}^{(\ell+1)} = \sin(\omega_0\cdot\mathbf{W}^{(\ell)}\mathbf{h}^{(\ell)}+\mathbf{b}^{(\ell)}), 
    ~\Phi(\mathbf{v}) = \mathbf{W}^{(L)}\mathbf{h}^{(L)}+\mathbf{b}^{(L)},
\end{equation}
where $\mathbf{W}^{(\ell)}\in\mathbb{R}^{d_{(\ell)}\times d_{(\ell+1)}},\mathbf{b}^{(\ell)}\in\mathbb{R}^{d_{(\ell+1)}}$ are layerwise parameters, and $\Phi(\mathbf{v})\in\mathbb{R}^{d_{\text{out}}}$ is the predicted value. $\sin(\cdot)$ is the periodic activation function with frequency factor $\omega_0$  \citep{SIREN}. $\gamma(\mathbf{v})$ is the concatenated random Fourier features (CRF) \citep{FourierFeature} with different Fourier basis frequencies $\mathbf{B}_k\in\mathbb{R}^{d/2\times c_{\text{in}}}$ sampled from the Gaussian $\mathcal{N}(0,\sigma_{k}^2)$:
\begin{equation}\label{eq:crf}
    \gamma(\mathbf{v})=[\sin(2\pi\mathbf{B}_1\mathbf{v}),\cos(2\pi\mathbf{B}_1\mathbf{v}),\dots,\sin(2\pi\mathbf{B}_{N_f}\mathbf{v}),\cos(2\pi\mathbf{B}_{N_f}\mathbf{v})]^{\mathsf{T}}\in\mathbb{R}^{d{N_f}}.
\end{equation}

By setting a large number of frequency features $N_f$ and a series of scale parameters $\{\sigma^2_k\}$, we can sample a variety of frequency patterns in the input domain.
The combination of these two strategies achieves high-frequency, low-dimensional regression, empowering the coordinate-based MLPs to learn complex details with high resolution.


\subsection{Factorizing spatial-temporal variability}
Using a single $\Phi_{\theta}$ to model entangled spatiotemporal interactions can be challenging. Therefore, we decompose the spatiotemporal process into two dimensions using variable separation:
\begin{equation}\label{eq:factorize}
    \Phi(\mathbf{v})=\Phi_{x}(v_x)\Phi_{t}(v_t)^{\mathsf{T}}, 
    \Phi_{x}:\mathcal{X}\mapsto\mathbb{R}, ~v_x\mapsto\Phi_{x}({v}_x)\in\mathbb{R}^{d_x}, ~\Phi_{t}:\mathcal{T}\mapsto\mathbb{R}, ~{v}_t\mapsto\Phi_{t}({v}_t)\in\mathbb{R}^{d_t}, 
\end{equation}
where $\Phi_{x}$ and $\Phi_{t}$ are defined by Eq. \eqref{eq:freqmlp}. Eq. \eqref{eq:factorize} is an implicit representation of matrix factorization model. But it can process data or functions that exist beyond the regular mesh grid of matrices. To further align the two components, we adopt a middle transform matrix $\mathbf{M}_{xt}\in\mathbb{R}^{d_x\times d_t}$ to model their interactions in the hidden manifold, which yields:
$\Phi(\mathbf{v}) = \Phi_{x}({v}_x)\mathbf{M}_{xt}\Phi_{t}({v}_t)^{\mathsf{T}}$.

\subsection{Generalizable representation with meta-learning}
Given a STTD instance, we can sample a set containing $M$ data pairs $\mathbf{x}=\{(\mathbf{v}_i,\mathbf{y}_i)\}_{i=1}^M$ where $\mathbf{v}_i\in\mathbb{R}^{c_{\text{in}}}$ is the input coordinate and $\mathbf{y}_i\in\mathbb{R}^{c_{\text{out}}}$ is the traffic state value. 
Then we can learn an INR using gradient descent over the loss $\min_{\theta}\mathcal{L}(\theta;\mathbf{x})=\frac{1}{M}\sum_{i=1}^M\Vert\mathbf{y}_i-\Phi_{\theta}(\mathbf{v}_i) \Vert_2^2$.
As can be seen, a single INR encodes a single data domain, but the learned INR cannot be generalized to represent other data instances and requires per-sample retraining.
Given a series of data instances $\mathcal{X}=\{\mathbf{x}^{(n)}\}_{n=1}^N$, we set a series of latent codes for each instance $\{\phi^{(n)}\in\mathbb{R}^{d_{\text{latent}}}\}_{n=1}^N$ to account for the instance-specific data pattern and make $\Phi_{\theta}$ a base network conditional on the latent code $\phi$ \citep{functa}. We then perform per-sample modulations to the middle INR layers:
\begin{equation}
    \mathbf{h}^{(\ell+1)} = \sin(\omega_0\cdot\mathbf{W}^{(\ell)}\mathbf{h}^{(\ell)}+\mathbf{b}^{(\ell)}+\mathbf{s}^{(n)}),~
    \mathbf{s}^{(n)} = h_{\omega}^{(\ell)}(\phi^{(n)})=\mathbf{W}^{(\ell)}_s\phi^{(n)}+\mathbf{b}^{(\ell)}_s,
\end{equation}
where $\mathbf{s}^{(n)}\in\mathbb{R}^{d_{(\ell)}}$ is the shift modulation of instance $n$ at layer $\ell$, and $h_{\omega}^{(\ell)}(\cdot|\omega\in\Theta):\mathbb{R}^{d_{\text{latent}}}\mapsto\mathbb{R}^{d_{(\ell)}}$ is a shared linear hypernetwork layer to map the latent code to layerwise modulations.
Then, the loss function of the generalizable implicit neural representations (GINRs) is given as:
\begin{equation}\label{eq:ginr_loss}
    \min_{\theta,\phi}\mathcal{L}(\theta,\{\phi^{(n)}\}_{n=1}^N;\mathcal{X})=\mathbb{E}_{\mathbf{x}\sim\mathcal{X}}[\mathcal{L}(\theta,\phi^{(n)};\mathbf{x}^{(n)}]=\frac{1}{NM}\sum_{n=1}^N\sum_{i=1}^M\Vert\mathbf{y}^{(n)}_i-\Phi_{\theta,h_{\omega}(\phi)}(\mathbf{v}_i^{(n)};\phi^{(n)}) \Vert_2^2.
\end{equation}

To learn all codes, we adopt the meta-learning strategy to achieve efficient adaptation and stable optimization.
Since conditional modulations $\mathbf{s}$ are processed as functions of $\phi$, and each $\phi$ represents an individual instance, we can \textit{implicitly} obtain these codes using an \textit{auto-decoding} mechanism.
For data $n$, this is achieved by an iterative gradient descent process: $\phi^{(n)}\leftarrow\phi^{(n)}-\alpha\nabla_{\phi^{(n)}}\mathcal{L}(\Phi_{\theta,h_{\omega}{(\phi)}},\{(\mathbf{v}_i^{(n)},\mathbf{y}_i^{(n)})\}_{i\in M})$, where $\alpha$ is the learning rate, and the above process is repeated in several steps.
To integrate the auto-decoding into the meta-learning procedure, inner-loop and outer-loop iterations are considered to alternatively update $\Phi_{\theta}$, and $\phi$. 

\section{RESULTS}
\begin{figure}[!htb]
\centering
\includegraphics[width=0.99\textwidth]{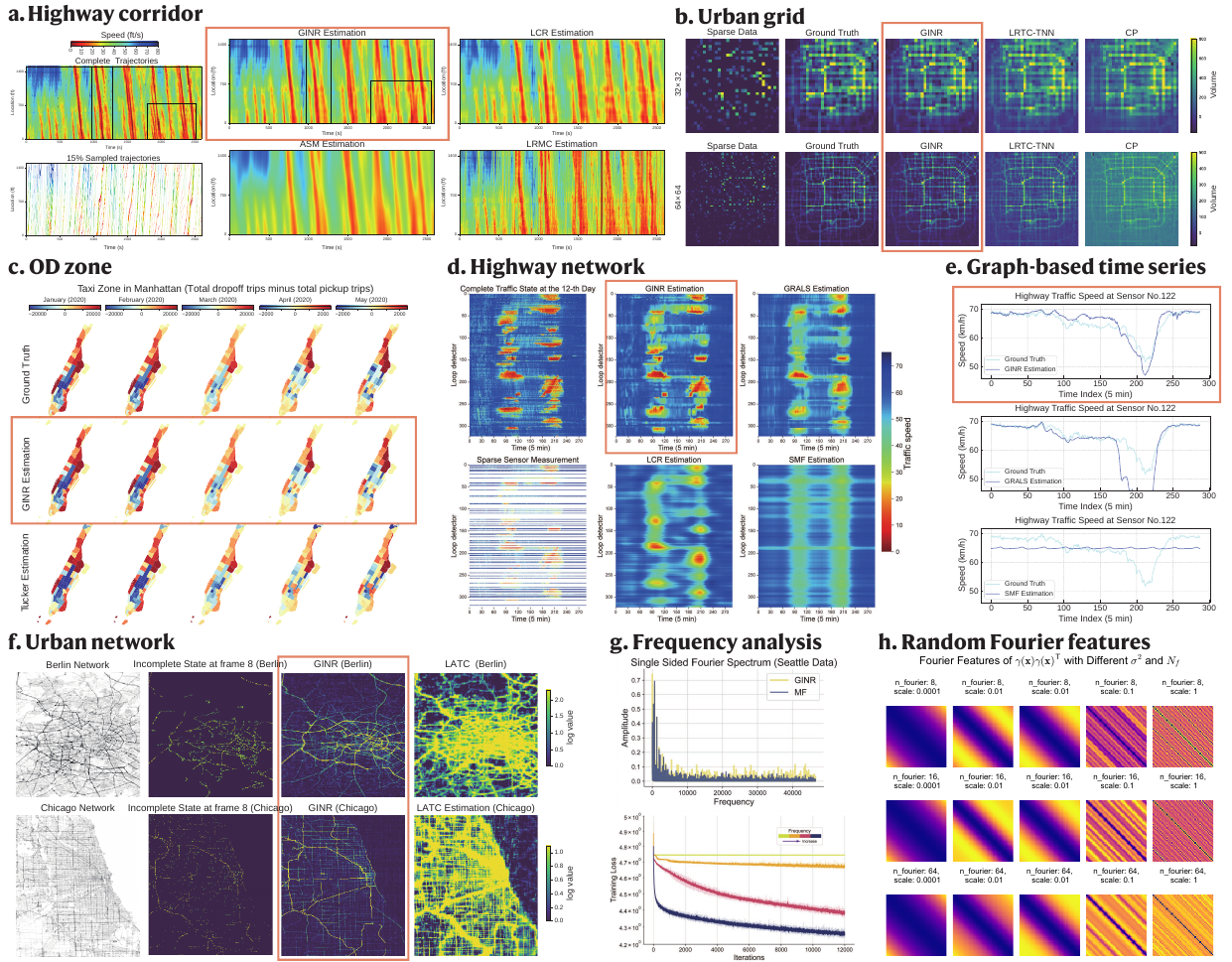}
\caption{Experiments on multiscale STTD. Full results can be found at \citep{nie2024spatiotemporal}.}
\label{fig:result_all}
\end{figure}

We conduct extensive experiments on real-world STTD covering scales from corridor to network, specifically including: (a) Corridor-level application: Highway traffic state estimation; (b-c) Grid-level application: Urban mesh-based flow estimation; and (d-f) Network-level application: Highway and urban network state estimation. We compare our model with SOTA low-rank models and evaluate its generalizability in different scenarios, such as different input domains, multiple resolutions, and distinct topologies. We also find that the encoding of high-frequency components is crucial for learning complex patterns (g-h). Fig. \ref{fig:result_all} briefly summarizes our results. 

\section{SUMMARY}
We have developed a new method for learning spatiotemporal traffic data (STTD) using implicit neural representations. This involves parameterizing STTD as deep neural networks, with INRs trained to map coordinates directly to traffic states. The versatility of this representation allows it to model various STTD types, including vehicle trajectories, origin-destination flows, grid flows, highway networks, and urban networks. Thanks to the meta-learning paradigm, this approach can be generalized to a range of data instances. Experimental results from various real-world benchmarks show that our model consistently surpasses conventional low-rank models. It also demonstrates potential for generalization across different data structures and problem contexts.

\begin{small}
\begin{sloppypar} 
\bibliographystyle{authordate1} 

\setlength{\bibsep}{0pt}

\bibliography{References}

\end{sloppypar}
\end{small}


\end{document}